# Quantum-enhanced long short-term memory with attention for spatial permeability prediction in oilfield reservoirs

Muzhen Zhang [1,†,*], Yujie Cheng [2,†,*], Zhanxiang Lei [1]

[1] Research Institute of Petroleum Exploration and Development (RIPED), PetroChina, Beijing, 100083, China

[2] Future Research Lab, China Mobile Research Institute, Beijing, 100053, China

[†] These authors contributed equally to this work.

[*] Correspondence: Muzhen Zhang (zhangmuzhen@petrochina.com.cn); Yujie Cheng (chengyujie@chinamobile.com)

**Abstract:**
Spatial prediction of reservoir parameters, especially permeability, is crucial for oil and gas exploration and development. However, the wide range and high variability of permeability prevent existing methods from providing reliable predictions. For the first time in subsurface spatial prediction, this study presents a quantum-enhanced long short-term memory with attention (QLSTMA) model that incorporates variational quantum circuits (VQCs) into the recurrent cell. Using quantum entanglement and superposition principles, the QLSTMA significantly improves the ability to predict complex geological parameters such as permeability. Two quantization structures, QLSTMA with Shared Gates (QLSTMA-SG) and with Independent Gates (QLSTMA-IG), are designed to investigate and evaluate the effects of quantum structure configurations and the number of qubits on model performance. Experimental results demonstrate that the 8-qubit QLSTMA-IG model significantly outperforms the traditional long short-term memory with attention (LSTMA), reducing Mean Absolute Error (MAE) by 19% and Root Mean Squared Error (RMSE) by 20%, with particularly strong performance in regions featuring complex well-logging data. These findings validate the potential of quantum–classical hybrid neural networks for reservoir prediction, indicating that increasing the number of qubits yields further accuracy gains despite the reliance on classical simulations. This study establishes a foundational framework for the eventual deployment of such models on real quantum hardware and their extension to broader applications in petroleum engineering and geoscience.

## 1. Introduction

Reservoir parameters are essential metrics for reservoir characterization and modeling, and their spatial prediction plays a critical role in oil and gas exploration and development. While well logging data provides the primary source for such predictions, limited drilling in actual field projects often hampers comprehensive reservoir understanding. Consequently, reliably forecasting reservoir parameters in undrilled areas using existing logging data remains a central challenge.

Machine learning techniques, with their strengths in modeling nonlinear relationships and handling high-dimensional complex data, have been widely applied to reservoir parameter prediction (Gamal and Elkatatny, 2022; Nguyen-Le and Shin, 2022; Nourani et al., 2022; Otchere et al., 2021; Wang and Cao, 2022; Wang and Cao, 2024; Zhou et al., 2025). Long short-term memory (LSTM) networks, which are particularly effective for time-series modeling, have been widely adopted to treat well-log depth variations as temporal sequences. These methods have been employed to generate or reconstruct missing logging curves from drilled wells using not only standard LSTM and its variants (Zhang et al., 2018), but also hybrid models that combine LSTM with other neural network architectures (Haritha et al., 2025; Shan et al., 2021; Wang et al., 2022; Zhang et al., 2023). The incorporation of attention mechanisms into LSTM further enhances feature extraction capabilities (Zhou et al., 2025). In undrilled areas, the LSTM-Attention (LSTMA) method has shown strong capabilities in predicting inter-well porosity logs (Zhang et al., 2024). However, permeability data—unlike porosity—have broader value ranges, greater variability, and more extreme values. These characteristics impose significantly higher demands on model accuracy, thereby rendering conventional neural network methods inadequate.

With ongoing advances in algorithms and the rapid development of quantum computing hardware, quantum machine learning (QML) has emerged as a major research focus (Biamonte et al., 2017). Quantum computing exploits superposition and entanglement to represent complex nonlinear structures in high-dimensional spaces with relatively few parameters. This capability offers potential advantages in accelerating optimization and improving convergence speed (Deutsch, 1985; Farhi et al., 2014; Feynman, 1982; Horodecki et al., 2009; Huang et al., 2022; Lloyd, 1996). The development of quantum neural networks (QNNs) (Beer et al., 2020) and quantum convolutional neural networks (QCNNs) (Wei et al., 2022) represents a deep convergence of classical deep learning and quantum computing. Previous studies have demonstrated that QNNs perform well in time-series forecasting tasks (Azevedo and Ferreira, 2007; Moon et al., 2025). In the financial domain, QNNs have been successfully applied to stock price prediction (Paquet and Soleymani, 2022; Gandhudi et al., 2024), with parametrized quantum circuit (PQC)-based QNNs even outperforming Bi-LSTM networks on highly noisy time-series data (Emmanoulopoulos and Dimoska, 2022). Building on classical LSTM networks, a quantum-classical hybrid architecture called Quantum LSTM (QLSTM) was proposed (Chen et al., 2022; Kea et al., 2024). QLSTM integrates a variational quantum circuit (VQC) into the basic LSTM cell to enable data exchange and transformation among gates (such as the input, output, and forget gates) via quantum circuit measurements. This model achieved accuracy comparable to classical LSTM in dynamical system forecasting while converging more rapidly.

Although QNNs have been employed in petroleum geoscience to capture weak seismic response features (Xue et al., 2021), classify reservoir fluids (Luo et al., 2024), recognize water-flooded layers (Zhao et al., 2019), assess $CO_2$ dissolution volumes (Rao et al., 2024), and interpret lithology from well logs (Liu et al., 2022), their application to reservoir parameter prediction remains limited. In particular, studies employing QLSTM for the spatial prediction of well-log curves are scarce. Existing LSTMA approaches also encounter difficulties when handling complex datasets such as permeability. Consequently, integrating QNNs into reservoir parameter prediction not only overcomes the modeling limitations of conventional neural networks but also offers a novel pathway for introducing quantum algorithms into petroleum geoscience applications.

To address these challenges, we propose an enhanced model, the Quantum Long Short-Term Memory with Attention (QLSTMA), which integrates quantum algorithms with neural networks to improve the spatial prediction of complex geological parameters, such as permeability. This model incorporates a quantum layer into the traditional LSTM architecture and employs a preprocessing approach that combines logarithmic transformation and normalization, effectively addressing issues such as uneven data distribution and sensitivity to extreme permeability values. The novel QLSTMA model significantly outperforms the conventional LSTMA model. Furthermore, we design and evaluate two distinct quantization structures, QLSTMA-SG and QLSTMA-IG, to investigate the effects of different quantum configurations and the number of qubits on predictive performance. Experimental results show that independent quantum layers (QLSTMA-IG), by enabling quantum optimization through separate information-processing pathways, provide more stable optimization compared to the shared quantum layer structure (QLSTMA-SG). Moreover, using the QLSTMA-IG model as an example, both predictive performance and convergence speed progressively improve as the qubit count increases from 4 to 6 to 8. The QLSTMA-IG model with eight qubits achieves superior accuracy and generalization in spatial permeability prediction, confirming its strong potential under complex geological conditions.

The remainder of this paper is organized as follows. Section 2 introduces the theoretical foundations and methods related to the proposed QLSTMA model. Section 3 describes the data preprocessing procedures and model construction process. Section 4 discusses the prediction performance of the model, with a focus on the impact of quantum structure design and the number of qubits. Section 5 discusses the main findings and provides the conclusions.

## 2. Principles and methods

### 2.1. Framework for spatial permeability prediction

We develop a framework for spatial permeability prediction by treating well-log curves as depth-wise sequences and applying a quantum-enhanced LSTM-Attention (QLSTMA) with adaptive distance and facies weighting. The central architecture of the proposed QLSTMA prediction method is a long short-term memory (LSTM) network that treats permeability logging curves from drilled wells as temporal sequences along depth $z_i$. By learning both the horizontal distribution and the vertical variation with depth, the model can invert these relationships to predict logging curves at undrilled locations. Each data point is represented by four equally treated temporal features $(x, y, z_i, f)$ and an associated permeability value $m_i, i = 1,2,...,n$, where $(x, y, z_i)$ are three-dimensional spatial coordinates and $f$ is a sedimentary microfacies code. At a given coordinate – facies combination $(x, y, f)$, there exist $n$ corresponding depth – permeability combinations $(z_i, m_i)$. During training, the LSTM simultaneously processes all four feature dimensions—selecting, retaining, forgetting, and transforming information—to build a model that estimates $m$ at any location within the same microfacies. Given fixed planar coordinates $(x_k, y_k, f_k)$ and the associated depth sequence $z_i$, the network predicts the corresponding permeability values $m_{ki}$, thereby reconstructing the logging curve shown in Fig. 1a. By design, the LSTM preserves continuity and correlation across these feature dimensions, making it particularly well suited for spatial prediction of well-log curves.

Sedimentary microfacies information is incorporated as a control variable during network training to reduce the discrepancy between predicted results and actual reservoir conditions, thereby enhancing prediction reliability. Consequently, the model requires both well coordinates $(x, y)$ and microfacies codes $f$ as inputs: the training weights assigned to each known well account for both spatial proximity and facies type.

For a prediction (undrilled) well, its permeability logging curve is predicted based on the logging data of drilled wells, with priority given to those closest in distance. Spatial weights are assigned inversely to well distance—closer wells carry greater weight. Facies weights reflect curve similarity: wells sharing the same sedimentary microfacies receive the highest weight; wells in other sedimentary microfacies are assigned weights in descending order based on their proximity in the facies sequence, from closest to furthest. In this study, sedimentary microfacies exhibit a cyclic lateral succession: distributary channel facies transition to inter-distributary sand facies, then to inter-distributary mud facies, before reverting to inter-distributary sand facies.

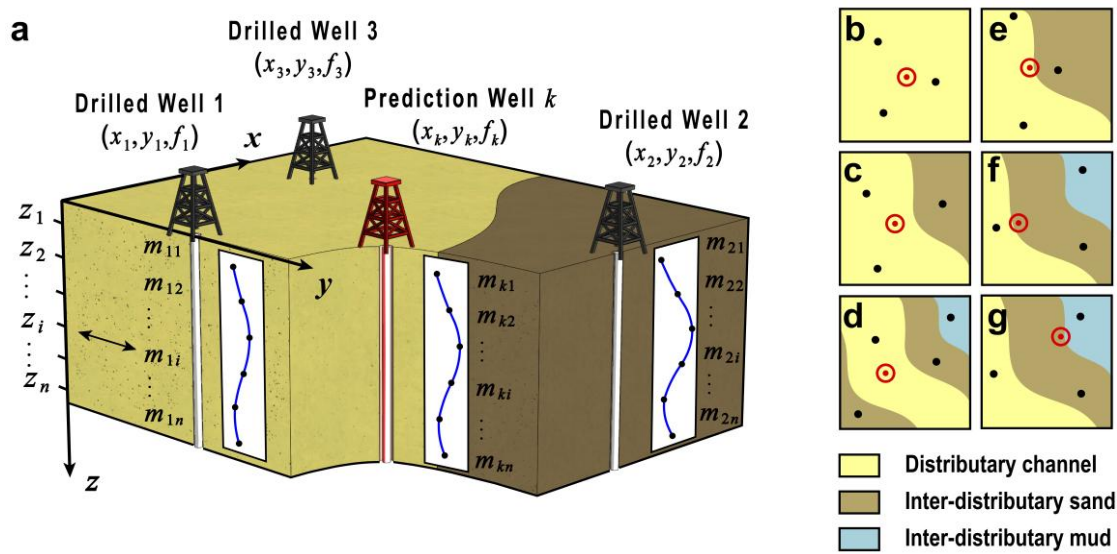

**Fig. 1.** Spatial permeability prediction framework with adaptive weighting. **a** Spatial permeability prediction using sequential log data from drilled wells. The schematic illustrates predicting permeability logging curves at an undrilled well using depth-wise sequences from surrounding drilled wells. Black wells denote drilled wells, red well denotes prediction (undrilled) well, and different colored regions represent distinct sedimentary microfacies. Each data point comprises spatial coordinates (x, y, z), sedimentary microfacies code (f) and permeability (m). **b–g** Adaptive weighting strategies combining spatial proximity and facies similarity, where data from drilled wells (black dots) and their weight assignments determine the spatial permeability prediction outcome of prediction wells (red-circled dots). Single microfacies weighting (**b**) is based solely on spatial proximity; multiple microfacies without conflicts (**c, d**) occur when spatial distance aligns with facies adjacency; conflict cases under multiple microfacies (**e, f, g**) require adaptive weighting that balances spatial proximity and facies similarity based on their relative importance.

When only distance is considered, the weighting of drilled wells is straightforward, as shown in Fig. 1b. Even when multiple sedimentary microfacies are present, distance and facies sequence order coincide, as in Fig. 1c and d. In practice, however, these factors often conflict. For example, a prediction well may lie very close to a drilled well yet belong to a different sedimentary microfacies, as shown in Fig. 1e, f, and g. Simultaneously weighting both well distance and sedimentary microfacies type introduces greater nonlinearity and complexity than predictions based solely on well coordinates, presenting a challenge for traditional methods. During prediction, interpolation must balance the influence of well distance and sedimentary microfacies type, adjusting each input data point's contribution to the model accordingly.

The QLSTMA model is a quantum neural network composed of a quantum-enhanced LSTM and an attention mechanism. The LSTM component learns from sequential data to capture the relationship between sedimentary microfacies variations and well-log curves. Quantization gives the LSTM greater expressive and generalization capacity. Specifically, classical LSTM units are replaced with parameterized variational quantum circuits (VQCs), which exploit quantum superposition and entanglement to capture nonlinear data features in a high-dimensional Hilbert space. Qubit interference further enables the model to learn long-range correlations among input features, enhancing its ability to model long-term dependencies. Moreover, incorporating VQCs substantially reduces the number of trainable parameters, allowing richer information to be represented with fewer parameters and improving the modeling of complex spatial correlations.

Additionally, the attention mechanism assesses the importance of various inputs and allocates higher weights to the most informative features. It prioritizes key aspects of data based on task requirements and data characteristics. Typically, QLSTMA assigns greater weight to wells nearer the prediction target. In areas of pronounced sedimentary microfacies variation, however, proximity alone may fail to reflect reservoir heterogeneity, so both well distance and facies characteristics must be considered. The attention mechanism dynamically adjusts these weights based on learned training patterns. For example, if it detects that well distance becomes less influential while facies type gains importance, it will shift the weighting toward sedimentary microfacies features. By optimizing this adaptive weighting, QLSTMA effectively accommodates complex depositional environments and irregular well distributions, ensuring robust prediction performance.

*2.2. Quantum LSTM architecture*

Based on the structural characteristics and limitations of LSTM networks, we integrate quantum computing principles into the standard LSTM framework to improve its expressive power and convergence speed in high-dimensional and strongly nonlinear settings. Accordingly, we propose two Quantum LSTM designs: QLSTM-SG (Shared Gate) and QLSTM-IG (Independent Gates). Their suitability and performance for spatial permeability prediction are evaluated in subsequent experiments.

QLSTM embeds trainable VQCs within each LSTM gating unit at every time step. By exploiting quantum superposition and entanglement, these circuits perform high-dimensional feature mapping and parallel information processing, thereby strengthening the model's capacity to capture complex dependencies.

*2.2.1. Variational Quantum Circuit (VQC) encoder*

Variational Quantum Circuits (VQCs) are a class of quantum circuit architectures for variational optimization and serve as a core component of the QLSTM model. By adopting a hybrid quantum–classical scheme, VQCs combine the expressive power of quantum computation with the flexibility of classical optimization. All key quantum operations—state preparation, entanglement, and measurement—are executed within the VQC, while loss evaluation and parameter updates based on measurement outcomes are performed classically. This configuration leverages the high-dimensional feature representation of quantum circuits while mitigating current hardware limitations in circuit depth and noise. In practice, the parameterized quantum circuit generates trial quantum states, and a classical variational optimizer iteratively adjusts the circuit parameters to maximize model performance.

Fig. 2 illustrates that the VQC within QLSTM comprises three stages. First, classical inputs are mapped to quantum states via Angle Embedding. Next, Strongly Entangling Layers introduce high-degree entanglement among qubits. Finally, Pauli-Z measurements and expectation value calculations yield the circuit's output. Each stage employs quantum gates to perform nonlinear mappings and propagate information. In Angle Embedding, the rotation gate $R_z$ maps classical data x = $[x_1, x_2, ..., x_n]$ to rotation angles that define the quantum state $|\psi\rangle$ as follows:

$$|\psi\rangle = \prod_{i=1}^{n} R_z(x_i)|0\rangle \quad (1)$$

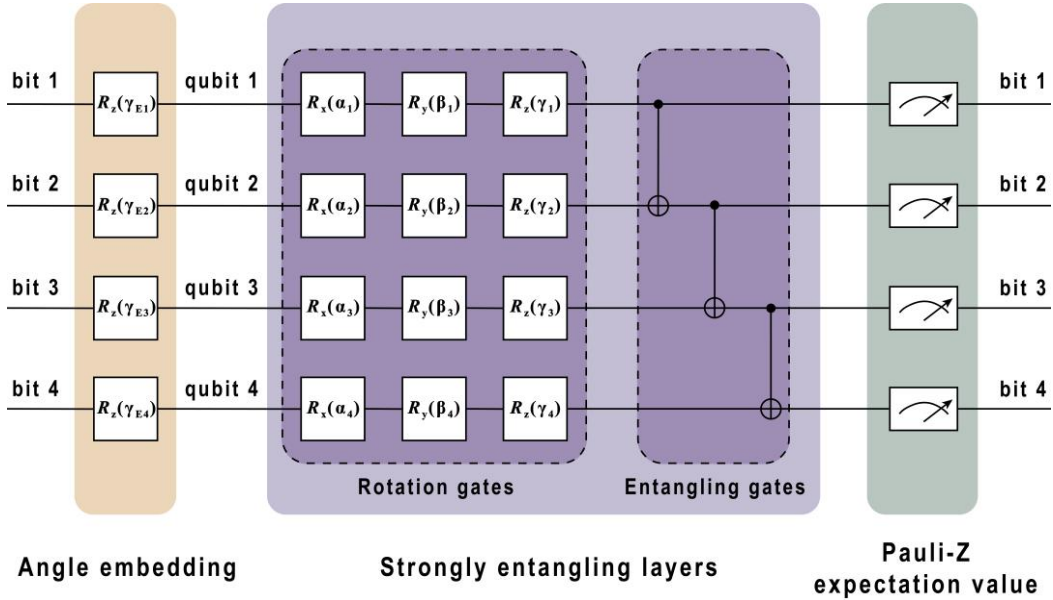

**Fig. 2.** VQC architecture for QLSTM. Classical bits are encoded into qubit states via Angle Embedding. In the Strongly Entangling Layers, rotation and entangling gates create and modulate multi-qubit entanglement to enable complex nonlinear interactions in the high-dimensional Hilbert space. Pauli-Z expectation values are then measured, converted back into classical features, and incorporated into the QLSTM architecture.

Subsequently, the Strongly Entangling Layers introduce trainable degrees of freedom through parameterized rotation gates and establish entanglement among qubits using controlled-NOT (CNOT) gates, thereby generating highly entangled quantum states $|\psi'\rangle$ in the high-dimensional Hilbert space and significantly boosting the circuit's expressive power. The mathematical form of a rotation gate acting on each qubit is given by the following equation:

$$|\psi'\rangle = \prod_{i=1}^{n_{layers}=1} \prod_{j=1}^{n_{qubits}} R_j(\theta_{ij})|\psi\rangle \quad (2)$$

Here, $R_j(\theta_{ij})$ denotes the rotation gate applied to qubit $j$, with the rotation angle $\theta_{ij}$ being a weight learned during training. In our implementation, $n_{layers} = 1$. These learned weights are stored in the VQC's Strongly Entangling Layers and are optimized alongside the other QLSTM parameters. Subsequently, CNOT gates are applied between adjacent qubits $i$ and $i + 1$ to establish entanglement. As shown in Fig. 2, the sequence consists of rotation gates on each qubit followed by CNOTs (i.e., entangling gates) on neighboring qubits to generate the desired entangled states.

At the end of the VQC, the Pauli-Z expectation value ⟨PauliZ⟩ is measured on each qubit to produce the final quantum outputs:

$$\langle PauliZ \rangle = \langle \psi' | \sigma_z | \psi' \rangle \quad (3)$$

Here, $\sigma_z$ denotes the Pauli-Z operator. The measured outputs are incorporated into the loss function, guiding optimization during subsequent training and prediction. During measurement, qubit superposition and entanglement collapse, converting quantum information into classical data. For an n-qubit system, the VQC produces an n-dimensional output vector. As one of the inputs to the LSTM, this vector contributes to computation and optimization and modulates the activation values of each LSTM gate.

In the VQC, parameterized rotation gates and fixed entangling gates, such as CNOT gates, are applied multiple times to construct the variational circuit. A classical optimizer then iteratively updates the rotation angles to minimize the target loss. As these angles are adjusted, the combined rotation–CNOT circuit flexibly explores the state space within the vast Hilbert space.

VQCs, by virtue of their parameterized architecture, can capture complex nonlinear relationships that are challenging for classical computers (Lubasch et al., 2020). The entanglement and superposition properties of qubits give VQCs a potential edge in addressing sophisticated tasks such as pattern recognition, optimization, and classification (Kwak et al., 2021). This hybrid approach combines quantum parallelism with classical optimization, mitigating current hardware limitations while harnessing quantum advantages. VQCs thus hold broad applicability and significant potential in fields ranging from quantum neural networks (Kwak et al., 2021) and quantum chemistry simulation (Shang et al., 2022) to quantum optimization problems (Le and Kekatos, 2024).

*2.2.2. Quantum LSTM variants: shared-gate (QLSTM-SG) and independent-gate (QLSTM-IG)*

This study proposes two QLSTM architectures: QLSTM-SG (Shared Gate) and QLSTM-IG (Independent Gates).

A conventional LSTM computes the forget gate $f_t$, input gate $i_t$, output gate $o_t$, and candidate cell state $\tilde{c}_t$ at each time step by applying four independent linear mappings (weight matrices plus biases) to the current input $x_t$ and the previous hidden state $h_{t-1}$. In the QLSTM-SG model, this classical linear transformation $W\begin{bmatrix}x_t\\h_{t-1}\end{bmatrix} + b$ is replaced by a VQC. The VQC first maps the concatenated inputs into quantum feature space, then performs a nonlinear transformation via its parameterized quantum gates. The circuit's output is fed into the usual activation functions to produce $f_t$, $i_t$, $o_t$, and $\tilde{c}_t$.

The computations of the three gates and the candidate state $\tilde{c}_t$ of the proposed QLSTM-SG neural network are expressed by the following equation:

$$\begin{bmatrix}f_t\\i_t\\\tilde{c}_t\\o_t\end{bmatrix} = \begin{bmatrix}\sigma\\\sigma\\\tanh\\\sigma\end{bmatrix}\left(\text{VQC}\left(W\begin{bmatrix}x_t\\h_{t-1}\end{bmatrix} + b\right)\right) \qquad (4)$$

Specifically, QLSTM resembles the classical LSTM by employing a recurrent unit at each time step that takes the current input $x_t$ and previous hidden state $h_{t-1}$ to produce the updated cell state $c_t$ and hidden state $h_t$. The key difference is that QLSTM embeds VQCs within this unit, creating quantum gating modules for the forget, input, and output gates as well as the candidate cell state. Unlike classical linear mappings, the VQC projects data into a high-dimensional Hilbert space, substantially enhancing the model's ability to represent complex nonlinear features.

In the QLSTM-SG model, all four gates share a single VQC. This configuration requires only one quantum circuit execution per time step: the same rotation angles and entangling layers produce pre-activation values for the forget, input, and output gates, as well as the candidate state, which are then passed through classical activation functions (such as sigmoid and tanh) to yield the final gate outputs. Although this design reduces computational cost, it merges all gating information within a single circuit, limiting quantization granularity and failing to fully exploit quantum superposition and entanglement to differentiate each gate's unique role. Consequently, QLSTM-SG represents an initial, suboptimal approach to LSTM quantization.

In the proposed QLSTM-IG architecture, each gate (forget $f_t$, input $i_t$, output $o_t$, and candidate cell $\tilde{c}_t$) is implemented with its own independent variational quantum circuit. Each circuit features dedicated rotation parameters and entangling layers, ensuring that the quantum mappings for one gate do not interfere with those of another. This higher degree of quantization, with per-gate quantum parameters, enables more precise extraction of the specific information required by each gate.

The computations of the three gates and the candidate state $\tilde{c}_t$ of the proposed QLSTM-IG neural network are expressed by the following equation:

$$\begin{bmatrix} f_t \\ i_t \\ \tilde{c}_t \\ o_t \end{bmatrix} = \begin{bmatrix} \sigma\left(\text{VQC}_1\left(W_f\begin{bmatrix} x_t \\ h_{t-1} \end{bmatrix} + b_f\right)\right) \\ \sigma\left(\text{VQC}_2\left(W_i\begin{bmatrix} x_t \\ h_{t-1} \end{bmatrix} + b_i\right)\right) \\ \tanh\left(\text{VQC}_3\left(W_c\begin{bmatrix} x_t \\ h_{t-1} \end{bmatrix} + b_c\right)\right) \\ \sigma\left(\text{VQC}_4\left(W_o\begin{bmatrix} x_t \\ h_{t-1} \end{bmatrix} + b_o\right)\right) \end{bmatrix} \quad (5)$$

Unlike the shared-gate scheme, QLSTM-IG implements each of its four gates with an independent VQC. Four separate parameter sets enable precise differentiation of each gate's quantum-state features, enhancing nonlinear fitting capacity, capturing complex temporal dependencies, and improving controllability. Although this design increases circuit calls and computational overhead, balancing predictive performance against resource consumption achieves superior results within acceptable cost limits.

*2.3. Attention-integrated hybrid LSTM and QLSTM architecture*

In the LSTMA model, combining LSTM with the attention mechanism creates a complementary framework. Initially, LSTM extracts fundamental long-term and short-term patterns from time-series data, where long-term patterns typically reflect extended geological evolution trends and short-term patterns indicate localized variations. Subsequently, the attention mechanism further emphasizes features critical to predictive accuracy. This integrated structure effectively captures both long-term trends and critical time steps, thereby better addressing the complexity and variability inherent in reservoir logging data for spatial prediction.

Integrating quantum neural network mechanisms into the LSTMA framework further enhances the model's ability to extract features of the spatial distribution of reservoir permeability. Subsurface reservoirs usually span large spatial areas, where long-range spatial dependencies significantly influence prediction accuracy. Quantum neural networks use quantum interference to preserve long-range spatial feature correlations, which enables the model to couple distant input variables via quantum phase. It further employs the linear superposition of quantum phase and amplitude to establish rich, complex feature correlations in a high-dimensional Hilbert space. This expressiveness endows QLSTMA with pronounced non-locality and long-range dependency modeling capabilities, thereby improving the precision of spatial permeability predictions.

Furthermore, quantum neural networks inherently exhibit implicit regularization, which enhances model generalization. Implicit regularization refers to the natural biases or constraints inherent in network architecture design, optimization algorithms, and training data preprocessing, which enable effective normalization of model parameters without introducing explicit penalty terms. QLSTMA's fixed quantum circuit structure and minimal parameter set generate strong implicit regularization, enabling rapid convergence and robust performance even on limited datasets.

Additionally, by using VQCs and quantum measurement mechanisms, quantum neural networks efficiently capture and represent the complex nonlinear relationship between input features and permeability. With far fewer trainable parameters than classical deep networks, QLSTMA provides expressive power and predictive accuracy comparable to or even surpassing its classical counterparts. This parameter efficiency not only substantially reduces the computational resources required for training but also enhances the model's ability to process complex geological information effectively.

## 3. Data preprocessing and model establishment

### 3.1. Datasets and preprocessing

This article presents a study of the U layer of an oilfield in Venezuela. The research area represents delta plain-to-front facies, characterized predominantly by distributary channel and inter-distributary bay sedimentary microfacies. Due to the presence of thin sand layers within inter-distributary bays, which differ significantly from the surrounding muddy layers in logging curves, the interdistributary bay facies was further subdivided into inter-distributary sand facies and inter-distributary mud facies. This classification facilitates improved feature extraction by the machine learning model based on distinct lithological characteristics.

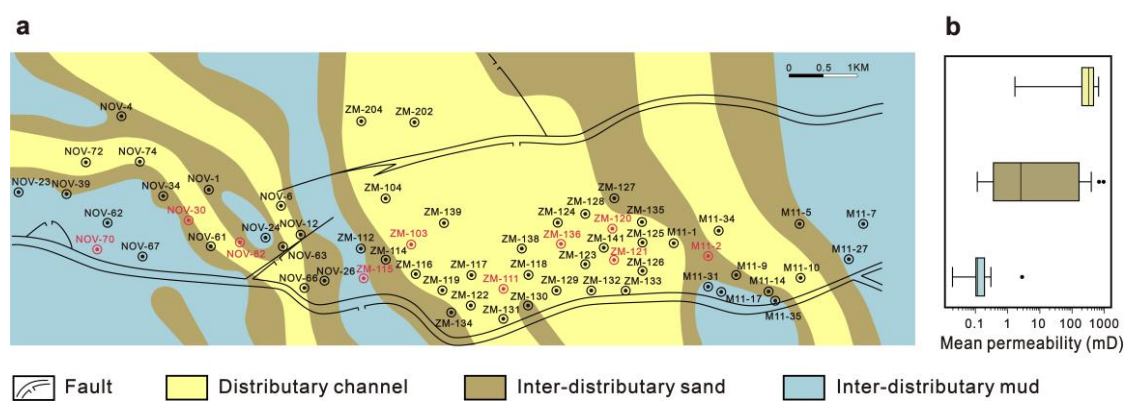

**Fig. 3.** Spatial distribution of sedimentary microfacies and permeability statistics in the U layer. **a** Spatial map of the U layer showing sedimentary microfacies and well locations. Black wells denote training wells, and red wells denote test (prediction) wells. **b** Boxplots of mean permeability for each well, grouped by sedimentary microfacies. In the boxplots, the center line denotes the median, box limits denote the upper and lower quartiles, whiskers denote the 1.5× interquartile range, and individual points denote outliers. Permeability (millidarcy) is plotted on a logarithmic axis.

A well-rarefying method was employed to validate the predictive capability of the trained model. The drilled wells were divided into training and testing datasets. During the training phase, test wells were excluded, and the model was constructed using only the data from the training wells. Subsequently, the model was applied to predict data from the test wells, and the predicted results were compared with the actual data to assess the model's performance. The U layer contains a total of 63 wells, including 53 training wells (in black) and 10 test wells (in red). The number of test wells for each sedimentary microfacies was proportionally allocated based on the distribution and total number of wells within each microfacies. Among the 63 wells, 34 training wells and 5 test wells belong to the distributary channel facies, 17 training wells and 3 test wells to the inter-distributary sand facies, and 12 training wells and 2 test wells to the inter-distributary mud facies. The sedimentary microfacies and well locations for the U layer are illustrated in Fig. 3a.

In this study, the porosity values range from 0 to 0.22, with individual well averages varying between 0 and 0.20. Permeability values span from 0.01 to 5026 mD, with individual well averages ranging from 0.02 to 953 mD. To describe the distribution of permeability, average permeability data from each well were statistically analyzed according to their respective sedimentary microfacies and visualized using boxplots, as shown in Fig. 3b. A logarithmic scale was used to clearly present data spanning several orders of magnitude.

Permeability values in distributary channel facies exhibit a relatively concentrated distribution, with a median permeability of 326.94 mD and an interquartile range (IQR) of 261.97 mD. Most permeability values fall between 100 and 1000 mD, indicating generally high permeability in channels. However, there are few data with lower permeability (approximately 1 mD), possibly due to localized variations in sedimentary conditions.

The permeability distribution within inter-distributary sand facies is notably wider, with a median permeability of 2.56 mD and an IQR of 164.37 mD, highlighting substantial permeability heterogeneity. Two high-permeability outliers are observed near 1000 mD, while the lowest measured value is approximately 0.1 mD. This wide variation reflects the transitional nature of this microfacies, situated between thick channel sands and muddy deposits, resulting in diverse sediment characteristics.

Inter-distributary mud facies show a narrower permeability distribution, with a median permeability of 0.12 mD and an IQR of 0.087 mD. Permeability values predominantly cluster in the low range, consistent with the muddy depositional characteristics of this microfacies. One notable higher permeability outlier (2.82 mD) was observed, possibly associated with localized sandy interbeds.

When employing neural network models to predict reservoir parameters, the difficulty varies across different types of parameters, primarily due to differences in their data ranges. A broader data range often corresponds to more complex logging curve details, thus increasing the prediction difficulty. Porosity, shale content, and water saturation typically use percentages as units, ranging from 0 to 1. Among these, porosity usually has a narrower and more concentrated range, often between 0 and 0.3. In contrast, permeability data frequently spans from 0 to thousands of millidarcies, occasionally reaching tens of thousands, featuring considerable variability and intricate internal details. Porosity data generally exhibit uniform distribution, whereas permeability data are often highly skewed with the presence of extreme values, requiring appropriate preprocessing techniques tailored to their characteristics.

Therefore, a logarithmic transformation is applied to permeability data. This transformation reduces data skewness, bringing the permeability data closer to a normal distribution, diminishing the influence of extreme values, and enhancing the effectiveness of statistical processing.

The logarithmic transformation is applied according to the following equation, with $c=1\times10^{-6}$, where $y_i$ denotes the input data and $y_{ln}$ the log-transformed data:

$$y_{ln} = \ln(y_i + c) \tag{6}$$

After the logarithmic transformation, further normalization processing is applied to ensure that all features are scaled consistently, facilitating better convergence and performance of machine learning algorithms. Additionally, as the logging data from each well varies in depth intervals and the number of sampling points, it is essential to standardize the input data before model training for consistency and comparability. A cubic spline interpolation method is employed to resample each well's data uniformly into 100 sampling points. Subsequently, input data normalization is performed, mapping the data onto a uniform range [0, 1]. After obtaining the model's predictions, an inverse normalization process is performed to recover the original depth and permeability values for each well.

The data normalization is applied according to the following equation, where $y^*$ denotes the normalized data and $y_{min}$ and $y_{max}$ denote the minimum and maximum values of the input data, respectively:

$$y^* = \frac{y_i - y_{min}}{y_{max} - y_{min}} \tag{7}$$

The selection of the loss function also affects the adaptability of neural network models to various data distributions. Typically, Mean Squared Error (MSE) loss is used in predictive models. However, considering the unique characteristics of permeability data, this study employs the Huber loss function instead of the MSE loss. The Huber loss is a hybrid function combining Mean Absolute Error (MAE) and MSE features, making it particularly suitable for data containing outliers. By introducing a threshold parameter $\delta$, the Huber loss applies a quadratic penalty (MSE) for smaller errors and a linear penalty (MAE) for larger errors. This design renders the loss function more robust to outliers while maintaining sensitivity to smaller errors.

The computation of the Huber loss is expressed by the following equation, where $y$ denotes the true value, $\hat{y}$ the predicted value, and $\delta$ the threshold parameter:

$$L_\delta(y,\hat{y}) = \begin{cases} \frac{1}{2}(y-\hat{y})^2, & \text{if } |y-\hat{y}| \leq \delta \\ \delta\left(|y-\hat{y}| - \frac{1}{2}\delta\right), & \text{otherwise} \end{cases} \tag{8}$$

*3.2. Quantum LSTMA model design for spatial permeability prediction*

We introduce two novel quantum LSTMA variants—QLSTMA-SG and QLSTMA-IG—using the classical LSTMA model as a baseline for comparison. The architecture of the LSTMA spatial permeability prediction model is illustrated in Fig. 4a and b. After logarithmic transformation and normalization, the input data enters the model through the input layer and is processed sequentially by the LSTM layer, attention layer, and time-distributed dense layer before being passed to the output layer. Each well is treated as one input sample with a data shape of (100, 4), where 100 corresponds to the number of sampling points (equivalent to time steps), and each time step contains four features: spatial coordinates $(x, y)$, depth sequence $z$, and the sedimentary microfacies code (0 for distributary channel facies, 1 for inter-distributary sand facies, and 2 for inter-distributary mud facies). The LSTM layer contains 64 hidden units. The attention layer generates weighted outputs using the Query, Key, and Value derived from the LSTM layer. The dense layer consists of 32 neurons and uses ReLU as the activation function. The model is optimized using the Adam optimizer and trained for 1000 epochs with dropout regularization (dropout rate = 0.5). The model's output is a predicted permeability logging curve corresponding to each input sample. Afterward, the output data undergo inverse normalization and inverse logarithmic transformation to restore them to their original scale, yielding prediction results suitable for interpretation and error analysis.

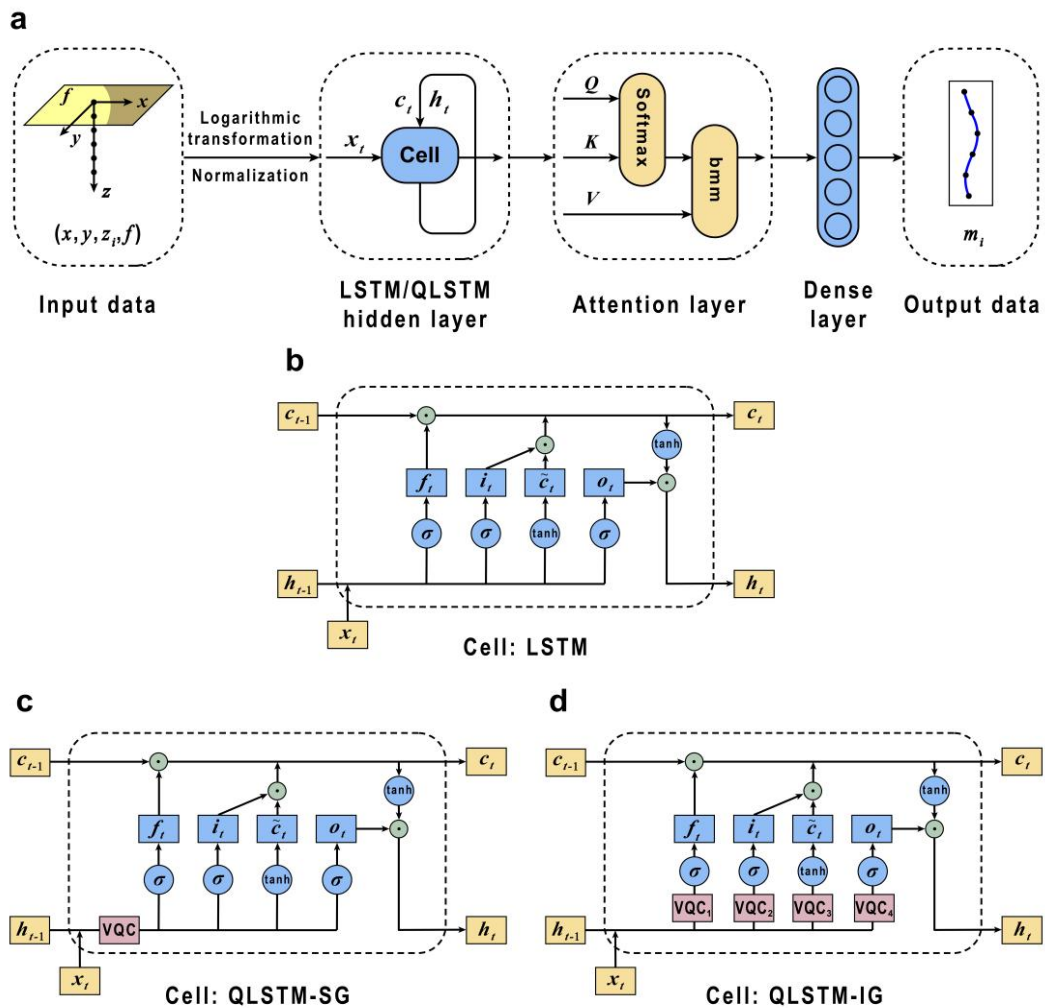

**Fig. 4.** Quantum-enhanced LSTM-Attention (QLSTMA) architectures for permeability prediction. **a** Overall neural network workflow includes logarithmic transformation and normalization of input data, LSTM layer, attention mechanism, dense, and output layers. **b** Classical LSTM cell utilizes traditional LSTM gates. **c** QLSTM-SG cell incorporates a shared-gate quantum variational circuit (VQC) module applied across all gates. **d** QLSTM-IG cell employs independent quantum variational circuits (VQCs) for each gate (forget, input, output, and candidate states), enabling more granular quantum feature extraction.

The QLSTMA-SG spatial permeability prediction model architecture is illustrated in Fig. 4a and c. "SG" refers to Shared Gate, indicating that all gates share a single quantum layer to extract quantum features. Except for the QLSTM hidden layer, the rest of the neural network architecture is identical to that of the LSTMA model. After the input data are fed into the LSTM layer, a variational quantum circuit (VQC) module is introduced into each LSTM recurrent cell. This module is applied after concatenating the current input $x_t$ with the previous hidden state $h_{t-1}$, and its quantum-transformed output is shared across the three gates and the candidate state. In QLSTM, the VQC performs the gate transformations that were originally handled by the classical linear mapping and activation function. Specifically, the linear mapping no longer directly produces the gate values. Instead, the outputs of the linear mapping are passed into the VQC for quantum feature mapping and nonlinear transformation, and an activation function then processes the resulting signals to obtain the final gate outputs. In QLSTMA-SG model training, the number of qubits is set to 8. All other model parameters remain the same as those in the LSTMA model.

The QLSTMA-IG spatial permeability prediction model architecture is illustrated in Fig. 4a and d. "IG" refers to Independent Gates, meaning that the forget gate, input gate, output gate, and candidate state each employ independent quantum layers for feature extraction. Except for the QLSTM hidden layer, the rest of the neural network architecture is identical to that of the LSTMA model. After the input data enters the LSTM layer, four VQC modules are embedded in each recurrent cell. These modules operate after the concatenation of the current input $x_t$ and the previous hidden state $h_{t-1}$. Unlike the QLSTMA-SG model, each gate and the candidate state in QLSTMA-IG are computed using separate quantum layers. During training, the number of qubits is set to 4, 6, and 8, respectively. All other model parameters remain the same as those in the LSTMA model.

## 4. Effects and analysis

### 4.1. Performance comparison of QLSTMA and LSTMA models

The five permeability prediction models, including LSTMA and the QLSTMA variants, were each trained independently five times for 1000 epochs. Model parameters were re-initialized before each run to reflect the randomness during model operation. All five models were trained and tested using the same input data, which included an identical set of training and testing wells. Each model generated predicted permeability logging curves for the same set of 10 test wells. The results from the five training runs of each model were averaged to obtain the final predicted permeability curve for each test well.

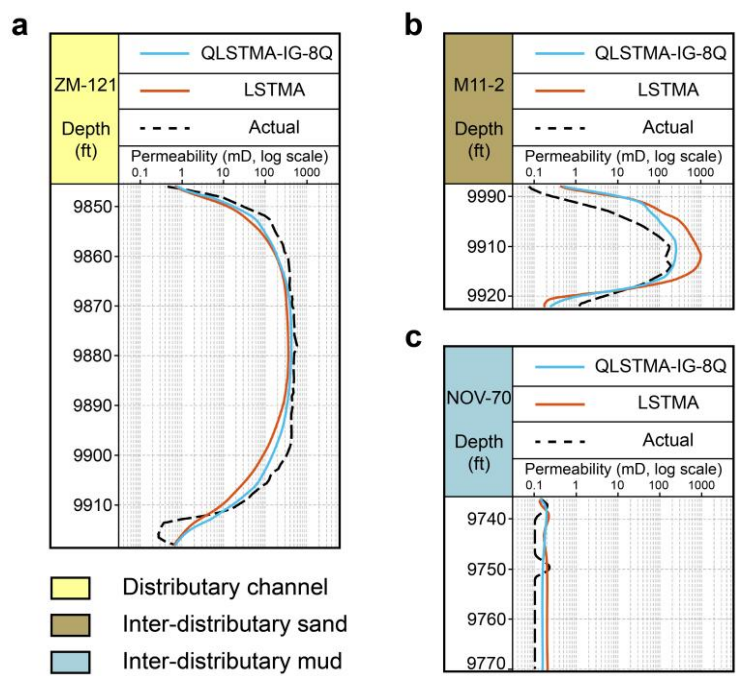

**Fig. 5.** Comparison of predicted and actual permeability logging curves for representative wells in three sedimentary microfacies. Model predictions from QLSTMA-IG-8Q and classical LSTMA are compared with actual measurements. **a** Well ZM-121 in distributary channel facies. **b** Well M11-2 in inter-distributary sand facies. **c** Well NOV-70 in inter-distributary mud facies.

Taking the QLSTMA-IG model with 8 qubits (QLSTMA-IG-8Q) as an example, Fig. 5 compares the permeability logging curves predicted by the QLSTMA and LSTMA models. One representative well is shown for each of the three sedimentary microfacies. In each subplot, the predicted curves from QLSTMA and LSTMA are shown in blue and red, respectively, while the black dashed line represents the actual logging curve. All plots are presented on a logarithmic scale.

As shown in the figure, both models can capture the general shape of the permeability logging curves across the three sedimentary microfacies. Notably, the predicted curves for the distributary channel and inter-distributary mud facies more closely resemble the actual curves than those for the inter-distributary sand facies. This can be attributed to the following reasons: the distributary channel facies has the largest number of training wells, with relatively concentrated spatial distribution, providing a solid data foundation for the model to learn its curve patterns. Although inter-distributary mud facies has fewer training wells, the permeability values are mostly close to zero, and the logging curves are highly similar and concentrated in distribution, which makes prediction relatively easier.

In contrast, the inter-distributary sand facies has fewer training wells that are more sparsely distributed. Additionally, this facies exhibits more complex sedimentary characteristics, leading to lower curve similarity among wells. Inter-distributary sand facies often forms under rapid sedimentation conditions. During fluvial processes, channel migration and swinging occur due to changes in external conditions, sediment supply, and dynamic forces. These processes cause abrupt variations in the composition and structure of sediments, resulting in significant heterogeneity in permeability and other reservoir properties in the inter-distributary sand facies.

The two models yield comparable results for the distributary channel and inter-distributary mud facies. However, for the more challenging inter-distributary sand facies, the QLSTMA model incorporating quantum circuits outperforms LSTMA. It produces predicted permeability curves that are closer in shape and magnitude to the actual measurements. For instance, LSTMA predictions reach a maximum of approximately 1000 mD, while both QLSTMA and the actual logging curve remain around 200 mD.

Mean Absolute Error (MAE) calculates the average absolute difference between predicted and true values. Since permeability data often exhibit significant variability, MAE provides a balanced assessment of prediction accuracy without being overly sensitive to outliers, making it suitable for macro-level analysis of permeability prediction results. Mean Squared Error (MSE), by squaring the prediction errors before averaging, gives greater weight to larger errors and is thus more sensitive to them. Root Mean Squared Error (RMSE), the square root of MSE, retains this sensitivity. In the context of permeability prediction, large localized errors may significantly impact the quality of the results. Therefore, this study uses both MAE and RMSE as error metrics to evaluate model performance, allowing a comprehensive assessment that considers both overall trends and local deviations in the predicted permeability curves.

The MAE and RMSE values were calculated between the average predicted results (from five independent runs) and the actual values for each test well across the five models, as shown in Table 1 and Table 2. Fig. 6 compare the MAE and RMSE of the five models across different sedimentary microfacies, based on the average error of test wells within each facies. Overall, models incorporating quantum algorithms exhibit lower error values and demonstrate notably better performance than the LSTMA model in inter-distributary sand facies, which are characterized by higher reservoir heterogeneity. This indicates that the incorporation of quantum algorithms with neural networks offers an advantage for permeability prediction. Furthermore, among the four QLSTMA models, those using the Independent Gates structure yield smaller prediction errors than those with the Shared Gate structure. In addition, models with a greater number of qubits tend to achieve better predictive performance.

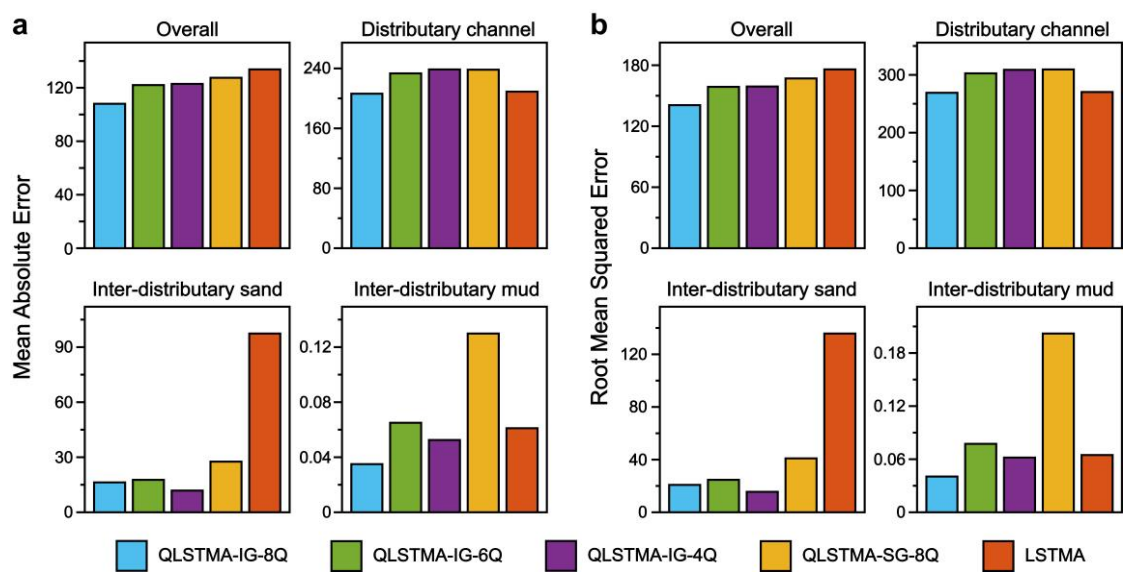

**Fig. 6.** Error comparison of permeability predictions across models and sedimentary microfacies. **a** MAE of permeability predictions by five models including QLSTMA-IG with 8, 6, and 4 qubits, QLSTMA-SG with 8 qubits, and classical LSTMA. Results are presented for all wells and separately for the three sedimentary microfacies of distributary channel, inter-distributary sand

and inter-distributary mud. **b** RMSE of the same models and facies subsets.

Table 1 Mean absolute error of permeability predictions across models and sedimentary microfacies.

| Prediction well | Sedimentary microfacies | QLSTMA-IG-8Q | QLSTMA-IG-6Q | QLSTMA-IG-4Q | QLSTMA-SG-8Q | LSTMA |
|---|---|---|---|---|---|---|
| ZM-111 | | 388.11 | 412.18 | 419.89 | 394.65 | 400.43 |
| ZM-120 | | 222.25 | 257.09 | 249.95 | 262.93 | 223.26 |
| ZM-121 | Distributary channel | 81.71 | 110.75 | 120.23 | 133.77 | 121.66 |
| ZM-136 | | 227.16 | 257.17 | 266.53 | 270.31 | 200.00 |
| ZM-103 | | 112.12 | 129.77 | 136.63 | 130.03 | 99.68 |
| Average | Distributary channel | 206.27 | 233.39 | 238.65 | 238.34 | 209.01 |
| M11-002 | | 48.59 | 52.75 | 35.20 | 82.55 | 291.97 |
| NOV-082 | Inter-distributary sand | 0.11 | 0.12 | 0.17 | 0.12 | 0.12 |
| NOV-030 | | 0.15 | 0.14 | 0.13 | 0.12 | 0.14 |
| Average | Inter-distributary sand | 16.28 | 17.67 | 11.83 | 27.60 | 97.41 |
| NOV-070 | Inter-distributary mud | 0.05 | 0.09 | 0.08 | 0.07 | 0.08 |
| ZM-115 | | 0.02 | 0.04 | 0.03 | 0.19 | 0.04 |
| Average | Inter-distributary mud | 0.03 | 0.07 | 0.05 | 0.13 | 0.06 |
| Overall average | | 108.03 | 122.01 | 122.88 | 127.47 | 133.74 |

Table 2 Root mean squared error of permeability predictions across models and sedimentary microfacies.

| Prediction well | Sedimentary microfacies | QLSTMA-IG-8Q | QLSTMA-IG-6Q | QLSTMA-IG-4Q | QLSTMA-SG-8Q | LSTMA |
|---|---|---|---|---|---|---|
| ZM-111 | | 505.45 | 538.75 | 546.48 | 515.99 | 521.17 |
| ZM-120 | | 252.39 | 293.22 | 283.34 | 299.71 | 252.01 |
| ZM-121 | Distributary channel | 100.22 | 132.28 | 142.65 | 157.03 | 145.13 |
| ZM-136 | | 315.11 | 341.39 | 353.45 | 359.27 | 281.98 |
| ZM-103 | | 171.75 | 207.62 | 218.02 | 215.01 | 151.38 |
| Average | Distributary channel | 268.99 | 302.65 | 308.79 | 309.40 | 270.33 |
| M11-002 | | 61.94 | 73.53 | 46.21 | 122.38 | 406.82 |
| NOV-082 | Inter-distributary sand | 0.18 | 0.18 | 0.28 | 0.21 | 0.21 |
| NOV-030 | | 0.30 | 0.23 | 0.18 | 0.24 | 0.29 |
| Average | Inter-distributary sand | 20.81 | 24.65 | 15.56 | 40.94 | 135.78 |
| NOV-070 | Inter-distributary mud | 0.06 | 0.09 | 0.08 | 0.07 | 0.09 |

| ZM-115 | | 0.02 | 0.06 | 0.04 | 0.33 | 0.04 |
|---|---|---|---|---|---|---|
| Average | Inter-distributary mud | 0.04 | 0.08 | 0.06 | 0.20 | 0.06 |
| Overall average | | 140.74 | 158.74 | 159.07 | 167.02 | 175.91 |

*4.2. Impact of quantum gate structure on QLSTMA performance*

For the LSTMA, QLSTMA-IG-8Q, and QLSTMA-SG-8Q models, training was conducted for 1000 epochs, and models were saved at every 10 epochs (i.e., epochs = 10, 20, ..., 1000). Based on these intermediate models' MAE and RMSE values, error evolution curves were plotted to visualize how the prediction errors evolved with training progress, as shown in Fig. 7a. Compared to QLSTMA, the LSTMA model exhibited a rapid error reduction in the early stage (before approximately 400 epochs), followed by a gradual increase in error with fluctuations, ultimately surpassing that of the QLSTMA models. This trend is especially evident in the RMSE curves, indicating that LSTMA is less capable of handling large localized deviations or sharp changes in permeability data. This observation is consistent with the findings that LSTMA performed significantly worse than QLSTMA in predicting inter-distributary sand facies, which are prone to sharp local variations and outliers.

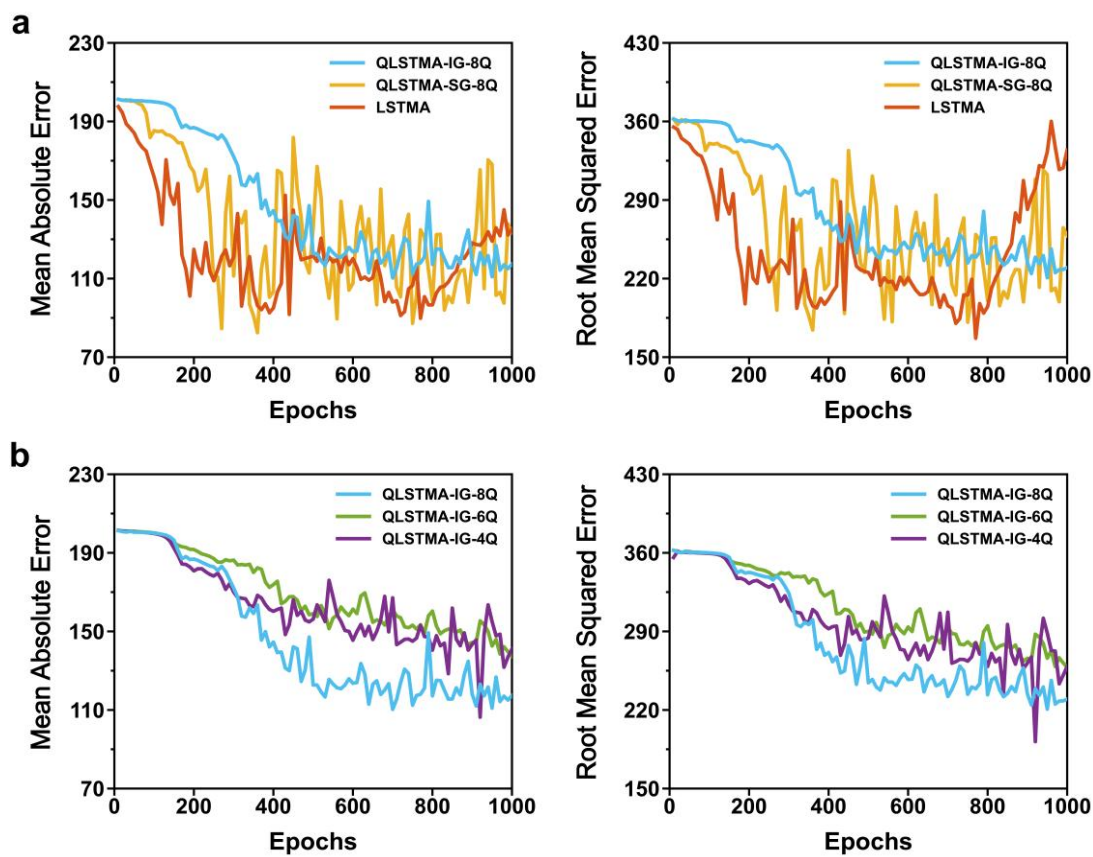

**Fig. 7.** Evolution of prediction error metrics during training across models and qubit configurations. **a** MAE and RMSE plotted against training epochs (evaluated every 10 epochs up to 1000) for classical LSTMA, QLSTMA-SG-8Q and QLSTMA-IG-8Q. **b** MAE and RMSE plotted against training epochs for QLSTMA-IG with 8 qubits, 6 qubits and 4 qubits.

The QLSTMA-SG model demonstrated a similar trend to LSTMA, with a fast initial decrease in error but a subsequent increase in the later stages of training. In contrast, the QLSTMA-IG model showed a more stable, steadily decreasing error curve throughout training, with fewer and smaller spikes. This comparison suggests that the independent quantization of the forget, input, and output gates, as well as the candidate state—implemented via four separate quantum layers—has a more positive impact on model performance than the shared quantum layer structure, which applies a unified transformation before the gating operations.

Although the Independent Gates structure leads to slower error reduction during the early training phase, it achieves more stable convergence in the later stages, reduces the risk of overfitting, and delivers better long-term performance. This phenomenon may occur because independently quantizing each gate allows for separate optimization of distinct information-processing paths within the model, enabling it to capture subtle feature differences in the data more effectively. In contrast, the shared quantum layer structure applies a unified transformation across all gates and thus reduces the total number of model parameters. However, this design overlooks the functional distinctions of individual quantum gates, resulting in less stable long-term prediction performance than the independent-gates structure.

*4.3. Impact of number of qubits on QLSTMA performance*

For the QLSTMA-IG-8Q, 6Q, and 4Q models, training was conducted for 1000 epochs, and models were saved every 10 epochs (i.e., epochs = 10, 20, ..., 1000). Based on these intermediate models' MAE and RMSE values, error evolution curves were plotted to visualize how the prediction errors evolved with training progress, as shown in Fig. 7b.

The three models using the Independent Gates structure exhibit a similar overall trend: a relatively smooth decline in error and smaller fluctuations during training. Among them, the model with 8 qubits demonstrates the most pronounced and stable reduction in error. Although the difference between the models with 6 and 4 qubits is less distinct than their difference from the 8-qubit model, the 4-qubit model shows sharper spikes and greater local fluctuations. It is also the least stable and yields the worst final prediction performance among the three.

These results indicate that, within the range of qubit numbers tested in this study, increasing the number of qubits leads to a corresponding improvement in the QLSTMA model's prediction performance. This improvement likely arises from the increased dimensionality of the Hilbert space enabled by more qubits, allowing the model to better capture non-linear and high-order feature interactions in permeability data.

**5. Discussion and Conclusions**

In this study, we introduced the QLSTMA model, a hybrid quantum–classical neural network integrating variational quantum circuits (VQCs) with long short-term memory (LSTM) and attention mechanisms to enhance spatial permeability predictions for oil and gas reservoirs. The QLSTMA framework significantly outperformed the classical LSTMA baseline, achieving reductions of approximately 19% and 20% in MAE and RMSE, respectively. Notably, performance improvements were most pronounced in regions with complex sedimentary features and high data heterogeneity, demonstrating the effectiveness of quantum algorithms in capturing intricate spatial dependencies.

Within the proposed quantization structures, the QLSTMA-IG (Independent Gates) variant exhibited superior performance over the QLSTMA-SG (Shared Gate) model, achieving further reductions of approximately 15% and 16% in MAE and RMSE, respectively. The Independent Gates structure allowed for separate quantum optimization of different information-processing paths within the model, effectively capturing subtle variations and enhancing model stability. Furthermore, our investigation revealed that, within the tested range, increasing the number of qubits progressively improved convergence speed and model stability. The 8-qubit QLSTMA-IG model demonstrated reductions of 11% in both MAE and RMSE compared to the 6-qubit model and 12% compared to the 4-qubit model. Increasing the number of qubits expanded the dimensionality of the Hilbert space, enabling the model to more effectively capture non-linear and high-order feature interactions in permeability data.

Integrating quantum algorithms endowed the QLSTMA model with implicit regularization properties, reducing overfitting risks and enhancing generalization capacity, which is particularly valuable in geological scenarios characterized by limited and noisy data. By utilizing fewer parameters relative to classical neural networks, the proposed quantum architecture provided both computational efficiency and robust performance, positioning it favorably for practical deployment on near-term quantum hardware. Although we simulated the quantum circuits on classical hardware, adopting a Noisy Intermediate-Scale Quantum ("Seeking a quantum advantage for machine learning", 2023) approach holds promise for near-term implementation on actual quantum computing platforms.

The practical implications of our study extend significantly into petroleum engineering. By improving permeability predictions in sparsely drilled regions, the QLSTMA framework supports enhanced reservoir characterization, modeling accuracy, and reserve assessments, contributing critical data for exploration and development planning. Moreover, its robust performance in noisy geological environments positions QLSTMA as a promising tool for real-time risk assessment and operational decision support throughout the exploration and production phases. These prospects underscore the significance of using quantum machine learning for subsurface spatial prediction in geoscience, marking an important step toward practical, domain-specific quantum advantage in applied sciences.

Looking forward, advances in quantum hardware and algorithms are expected to further amplify the capabilities demonstrated here. Quantum optimization methods may revolutionize reservoir development strategies, facilitating multi-factor optimization to enhance oil recovery efficiency. Quantum simulation techniques have already shown substantial promise for accelerated modeling of reservoir flow dynamics, potentially alleviating computational bottlenecks encountered in traditional numerical simulations (Pfeffer et al., 2023). Quantum communication technology, with its inherent security advantages, may underpin future digital management systems for intelligent oilfield operations.

**Nomenclature**

| | |
|---|---|
| $f_t$ | Output of the forget gate layer at time $t$ |
| $i_t$ | Output of the input gate layer at time $t$ |
| $o_t$ | Output of the output gate layer at time $t$ |
| $c_t$ | Internal state of the processor at time $t$ |
| $h_t$ | Hidden state of the processor at time $t$ |
| $\tilde{c}_t$ | Candidate state of the processor at time $t$ |
| $\sigma$ | Logistic function |
| $W, b$ | Weights and bias terms of the corresponding layer |
| $x_t$ | Input at time $t$ |
| $\lvert \psi \rangle$ | Quantum state |
| $\sigma_z$ | Pauli Z operator |
| $x_i$ | Classical input feature $i$ for angle embedding |
| $R_z(x_i)$ | Single-qubit rotation gate about the $z$ axis with angle $x_i$ |
| $n$ | Number of qubits (dimension of embedding) |
| $n_{layers}$ | Number of variational-circuit layers |
| $\theta_{ij}$ | Rotation-angle parameter applied to qubit $j$ in layer $i$ |
| $\lvert \psi' \rangle$ | Quantum state after all variational layers |
| $\langle PauliZ \rangle$ | Expectation value of $\sigma_z$ on state $\lvert \psi' \rangle$: $\langle \psi' \lvert \sigma_z \rvert \psi' \rangle$ |
| $y_i$ | Input data |
| $y_{ln}$ | Log-transformed data |
| $y^*$ | Normalized data |
| $y_{min}$ | Minimum value of input data |
| $y_{max}$ | Maximum value of input data |
| $y$ | True value |
| $\hat{y}$ | Predicted value |